\documentclass[10pt, a4paper]{article}
\usepackage{lrec2022} 
\usepackage{multibib}
\newcites{languageresource}{Language Resources}
\usepackage{graphicx}
\usepackage{tabularx}
\usepackage{soul}
\usepackage{xcolor}
\usepackage{titlesec}
\titleformat{\section}{\normalfont\large\bfseries\center}{\thesection.}{1em}{}
\titleformat{\subsection}{\normalfont\SmallTitleFont\bfseries\raggedright}{\thesubsection.}{1em}{}
\titleformat{\subsubsection}{\normalfont\normalsize\bfseries\raggedright}{\thesubsubsection.}{1em}{}
\renewcommand\thesection{\arabic{section}}
\renewcommand\thesubsection{\thesection.\arabic{subsection}}
\renewcommand\thesubsubsection{\thesubsection.\arabic{subsubsection}}

\usepackage{epstopdf}
\usepackage[utf8]{inputenc}

\usepackage{hyperref}
\usepackage{xstring}

\usepackage{color}

\title{DialCrowd 2.0: A Quality-Focused Dialog System Crowdsourcing Toolkit\\ \vspace*{.5\baselineskip}}

\name{Jessica Huynh, Ting-Rui Chiang, Jeffrey Bigham, Maxine Eskenazi} 

\address{Carnegie Mellon University \\
         Pittsburgh, PA \\
         \{jhuynh, tingruic\}@cs.cmu.edu, \{jbigham, max\}@cmu.edu}

\abstract{
Dialog system developers need high-quality data to train, fine-tune and assess their systems. They often use crowdsourcing for this since it provides large quantities of data from many workers. However, the data may not be of sufficiently good quality. This can be due to the way that the requester presents a task and how they interact with the workers. This paper introduces DialCrowd 2.0 to help requesters obtain higher quality data by, for example, presenting tasks more clearly and facilitating effective communication with workers. DialCrowd 2.0 guides developers in creating improved Human Intelligence Tasks (HITs) and is directly applicable to the workflows used currently by developers and researchers. 
\newline  }

\begin{document}

\maketitleabstract

\section{Introduction}
High-quality human data is essential in the development of dialog systems. Many researchers create HITs on crowdsourcing platforms such as Amazon Mechanical Turk (AMT) to collect data from humans. Obtaining high-quality data is dependent on the usability of the tasks workers are asked to complete (\textit{e.g.}, learnability, feedback, etc.) \cite{nielsen1994usability}, yet many tasks fall short \cite{huynh2021survey}.


To address this problem, we introduce DialCrowd 2.0, a substantial update to DialCrowd 1.0. DialCrowd 1.0 \cite{lee2018dialcrowd} facilitated data collection by providing an interface that requesters used to create HITs from pre-configured templates. The goal in the 1.0 version was to make the task creation process more efficient. Once a HIT was created, workers accessed and worked on the HIT from the DialCrowd-generated interface. The added efficacy that DialCrowd provides was studied with 10 participant/requesters. All participants observed that DialCrowd shortened the time spent creating the study, and when asked to rate the usefulness of this toolkit, participants responded with an average of 4 on a scale of 1 to 5, with 5 being best.

Whereas DialCrowd 1.0 focused on helping requesters create HITs more efficiently, DialCrowd 2.0 addresses factors related to interaction and communication with the workers that can affect the quality of the data obtained from a HIT.
%
We have demonstrated the community's need for help with these two aspects in a recent study
\cite{huynh2021survey}. In this study, we looked at the tasks running on AMT over seven consecutive days in August 2021 to analyse their overall quality. The study examined only the natural language processing HITs (excluding computer vision, surveys, etc.) that were presented to workers at this time. In the same way that requesters can give ratings to an individual worker, workers also rate the requesters and share information about them on their crowdsourcing forums and blogs \cite{paolacci2010running}. A high requester rating will attract more good workers while poor ratings and issues in communication with the workers will repel the better workers. Thus, in parallel to our examination of the HITs on AMT, we also tallied the workers' assessments of these HITs and of their requesters using Turkerview \footnote{\url{https://turkerview.com}}. Out of a total of 102 HITs available over that time span, 56 met our criteria and were reviewed for the study. 54 of the total 102 HITs were reviewed on Turkerview for payment and 67 out of a total of 79 requesters were reviewed on Turkerview for payment assessment.

Reinforcing the hypothesis that requesters need help with their HITs, we found that 25\% of the 56 HITs had technical issues. Out of the 54 HITs reviewed on Turkerview, only 39\% paid above \$10 an hour. All of the payment levels may be found in Figure \ref{table:payment_amt} \cite{huynh2021survey}. The findings in this study reinforce the claim of this paper that the research community needs DialCrowd 2.0 to help them obtain better quality crowdsourced data.

\begin{table}[h!]
\centering
\begin{tabular}{|c|c|c|} 
 \hline 
 Payment & No. HITs & \% of HITs \\
 \hline \hline
 $<$ \$7.25 & 24 & 44\% \\
 \hline
 \$7.25 - \$10.00 &  9 & 17\% \\
 \hline
 $>$ \$10.00 & 21 & 39\% \\
 \hline
\end{tabular}
\caption{Payment Statistics for HITs}
\label{table:payment_amt}
\end{table}

\section{Related Work}

\subsection{Dialog Data Collection}
Tools such as ParlAI \cite{miller2017parlai}, ConvLab \cite{zhu-etal-2020-convlab}, and MEEP \cite{arkhangorodsky2020meep} were created to make HIT creation easier. ParlAI and ConvLab are directly integrated with AMT with some coding required. MEEP is not integrated with AMT, but has a Wizard-of-Oz interface for data collection. In all three cases, these tools focus on providing pretrained models, datasets, and instruction on dialog system creation. However, they do not provide a guide to communication and clarity with workers during HIT creation, which DialCrowd 2.0 does offer.

\subsection{High Quality HITs}
We define a high quality HIT as being a HIT that both gathers high quality data and one that affords better quality communication and respect between requesters and workers. The worker wants to do the task correctly while minimizing the amount of time they spend on it (thus maximizing the amount they are paid per hour). Thus they will choose to work on the task that enables them to maintain this balance the best  \cite{faradani2011s}. The requester, on the other hand, wants to gather and aggregate many workers' responses in order to produce good quality data to train or assess their dialog system or for a study \cite{wang2017cost}. 

While the requester is rating the workers and choosing workers with a high rating to do their HIT, the workers are also rating the requesters in order to choose whose HIT to work on. A high requester rating attracts more good workers while poor ratings and issues in communication with the workers repel the better workers. Our above-mentioned survey found that 35\% of the 67 requesters studied were judged by workers as paying very badly or poorly \cite{huynh2021survey}.

This paper defines and implements five criteria that DialCrowd 2.0 incorporates to contribute to a high quality production: include clear instructions and examples, allow workers to provide feedback, pay workers fairly, filter out low quality work, and filter outlier data.

\subsubsection{Providing Instructions}
The first thing workers see when accessing a HIT is the set of instructions. The requester can improve the task and attract the better workers by giving a high level description of what the data will be used for and by providing clear and unambiguous instructions about what to do  \cite{chandler2013risks}. Requesters can also improve the interactive aspects of the interface the worker sees so less time is spent scrolling and searching \cite{marcus2012counting} \cite{daniel2018quality}. Chen et al 2011 \cite{chen2011opportunities} and Georgescu et al 2012 \cite{georgescu2012map} have shown that attending to interactive issues improves data quality.

Our above-mentioned study \cite{huynh2021survey} found that 28\% of the 56 HITs had incomplete, unrelated, or ambiguous instructions. More detail is shown in Figure \ref{table:instr_issues}.

\begin{table}[h!]
\centering
\begin{tabular}{|c|c|c|} 
 \hline 
 Instr. Issue & No. HITs & \% of HITs \\
 \hline \hline
 Completely Unclear & 0 & 0\% \\
 \hline
 Incomplete & 12 & 22\% \\
 \hline
 Unrelated & 2 & 4\% \\
 \hline
 Ambiguous/Vague & 1 & 2\% \\
 \hline
\end{tabular}
\caption{Instruction Issues}
\label{table:instr_issues}
\end{table}

When presented with ambiguous instructions, workers may rely on their previous experience with similar tasks to create their own interpretations of what they are to do \cite{chandler2013risks}. To improve this aspect of the  instructions, TaskMate has workers discover ambiguities in the instructions before the entire task is released \cite{k2019taskmate}. An automatic model that evaluates the instructions may also help a requester see how clear their instructions are \cite{nouri2021unclear}.


\subsubsection{Providing Examples}
 The use of well-chosen examples and counterexamples with accompanying explanations of why these particular examples were presented also helps workers to better understand the task. Providing these examples has been shown to improve data quality over other methods such as using gold standard questions \cite{doroudi2016toward}.

\subsubsection{Feedback}
Another way to improve communication with the workers is to give them a text box at the end of each task where they can provide feedback \cite{kittur2013future}. One study created a feedback drop-down menu that gives workers a list of specific reason for the feedback. While this is more restricted, it does allow the worker to pinpoint potential issues in the HIT more rapidly \cite{kulkarni2012mobileworks}. The use of a menu has not been shown to be correlated with an immediate increase in data quality.

\subsubsection{Fair Payment}
It is important to pay workers fairly for their time and effort. There are conflicting studies on whether higher payment levels increase the quality of data. Some studies show significant increases in data quality \cite{aker2012assessing}, some show that data quality increases up to a certain amount and then starts to decrease \cite{feng2009acquiring}, while others show that data quality stays the same but that the speed at which the HIT is finished is faster when payment is lower \cite{mason2009financial} \cite{buhrmester2016amazon} \cite{paolacci2010running}. DialCrowd underlines the importance of paying the workers a minimum wage of \$15/hr.

\subsubsection{Identifying Low Quality}
The filter most frequently used for low quality data detection has been gold standard HITs (HITs that have previously been completed by the requester or some expert) \cite{alabduljabbar2019dynamic}. This data is used to check whether the worker's production agrees with that of the expert \cite{allahbakhsh2013quality} \cite{chen2011opportunities} \cite{hsueh2009data} \cite{sayeed2011crowdsourcing} \cite{daniel2018quality}. These gold standard HITs have been shown to have benefits beyond just assessing one worker's production. They can also be used to find consistent bias, or imbalanced datasets \cite{wang2011managing}. Another filter uses duplicated data \cite{alabduljabbar2019dynamic}. In this case the requester has a worker do the same HIT twice during the course of their work. The hope is that the worker will give the same answer both times, thus demonstrating intra-worker consistency. Both of these methods are, evidently, not cost efficient since requesters are asking for duplicate work, but they do help improve quality.   

\subsubsection{Identifying Outliers}
Yet another option is to filter the data gathered for outliers. This includes pattern matching (for example, if a worker has selected answer choice A for every question), in order to measure an individual worker's reliability and agreement with the rest of the workers' output \cite{chandler2013risks} \cite{daniel2018quality}, as well as the amount of time spent \cite{rzeszotarski2012crowdscape}.

\section{DialCrowd 2.0}
Using what is known about best crowdsourcing practices, DialCrowd 2.0 helps requesters create HITs according to those practices. This section presents DialCrowd 2.0, which can be accessed at the following link: \url{https://cmu-dialcrowd.herokuapp.com/}.

\subsection{Task Creation}

DialCrowd 2.0 has a user-friendly interface that helps requesters to create tasks more easily. After consulting many publications that use crowdsourcing, four types of tasks stood out as being the most often used. Thus task templates were created for these four task types and more templates can be added by the DialCrowd team upon request:
\begin{itemize}
    \item Interactive task: workers interact with a dialog agent. This template can be used to collect conversation with dialog agents for training or to assess dialog agents.
    \item Intent classification: workers classify the intent of an utterance.
    \item Entity classification: workers label the entities in an utterance.
    \item Quality annotation: workers assess the quality of a dialog system's response given a context and response pair.
\end{itemize}

Requesters use one of the templates and then only need to fill out predefined configuration fields using DialCrowd 2.0's web-based graphical user interface to create a task. This eliminates the need to manually edit HTML code. Other related minor features are also provided as seen in Figures 1, 2, 3, and 4 in the Appendix, which show some examples of what the configuration page looks like. Figures 5 and 6 show what the workers see.

\begin{itemize}
    \item Serializable configuration: Requesters can upload and download task configuration files in JSON format. It helps requesters duplicate tasks or generate tasks automatically with programs. 
    \item Flexible appearance: DialCrowd 2.0 supports Markdown, which is a lightweight mark up language. It helps requesters format text easily. DialCrowd 2.0 also allows requesters to customize the style of a task, e.g. background color, text font.
    \item Calculation of worker payment: While this is not a minor issue, it is dealt with in a succinct and efficient manner. The requester has several persons work on the given task and determines the average amount of time it has taken them to accomplish the task. That amount is entered and DialCrowd 2.0 uses this number to suggest worker payment, based on an hourly wage of 15 dollars an hour.
    \item Calculation of the number of tasks to deploy: DialCrowd 2.0 calculates the number of tasks to deploy on AMT based on the data the requester has uploaded, the number of items/assignments per task unit, and the number of task units per task.
    \item Built-in consent form upload: DialCrowd 2.0 has a built-in function for adding consent forms and their corresponding check boxes.
\end{itemize}

\subsubsection{Clarity} 
Instructions that are clear and unambiguous help maintain better bidirectional communication between the requester and the workers. While the requesters create clear instructions, the workers give feedback on how to make the HIT better. It is good practice to post a small subset the total HITs first. In this way resulting quality can be assessed and feedback can be gathered from the workers. This allows for improvements to be made in the task before it is completely deployed and avoids the high cost of needing to repost a whole HIT when the resulting data has been poor.  

For requester-to-workers communication, DialCrowd 2.0 gives requesters guidance on how to compose clear and complete instructions on the DialCrowd 2.0 configuration page. There is also a link to the AMT best practices guide. DialCrowd 2.0 also explains the importance of giving examples and counter examples and provides space for requesters to input these items along with explanations of why both types of examples were chosen.  

For worker feedback, DialCrowd 2.0 includes an optional feedback space which gives workers the opportunity to point out instructions that are hard to follow, suggest better layout, note something that is not functioning correctly etc. While the abovementioned practice of posting a small amount of tasks first may seem counterintuitive and one might wonder if workers will actually take the time to fill out an optional text box if they are not paid more, \cite{mortensen2017exploration} showed that workers do indeed provide feedback.

\subsubsection {Low-Quality Data Detection} 

Even a well-constructed task may yield some low quality work. This may be due to the work of bots, carelessness or fatigue on the part of a worker. For this, DialCrowd 2.0 provides detection analytics that include quality control tasks and metrics for anomaly detection. It should be noted that the longer a HIT is active, the more likely it is that there will be bots working on it.

DialCrowd 2.0 offers two types of quality control tasks. (1) it helps requesters include \emph{duplicated tasks}, which can be used to check individual worker consistency (intra-worker agreement). As mentioned above, the data in a HIT is shown twice to a worker at different places. A consistent worker is expected to complete the same HIT in the same way both times they see it. (2) DialCrowd 2.0 also enables requesters to upload \textit{golden data} as described above. The worker's output is compared to the experts' and data that does not match can be eliminated. If a given worker's output frequently does not match that of the expert, the totality of that worker's data may be eliminated (but the worker should still be paid for the time they spent trying to do the task).

DialCrowd 2.0 also helps requesters detect worker behavior that differs from other workers with the following metrics: 
\begin{itemize}
    \item Time: DialCrowd 2.0 tracks the amount of time spent by each worker on the task. DialCrowd 2.0 flags work that is two standard deviations away from the mean time taken by all of the other workers to accomplish the task. A very short period of time, for example, may indicate the presence of a bot, while a very long period of time may indicate unfamiliarity with the goal or the content of the task.
    \item Patterns: A worker's answers may reveal a pattern in multiple choice answers. Responding A to every question, is an example of data that DialCrowd 2.0 will flag, thus providing another way to detect potential bots.
    \item Agreement: For inter-worker agreement, DialCrowd 2.0 calculates the agreement between each worker and all the other workers on the same HIT using Cohen's Kappa.
\end{itemize}

For each task, DialCrowd 2.0 provides a data summary page with all of the above information. This includes a table breaking down the summary numbers into individual results of these quality checks. It also includes individual Cohen's Kappas between raters for each of the questions asked, as well as the Cohen's Kappa among raters for all of the questions as a whole.

\section{Observations}
Although DialCrowd 2.0 provides guidance for many aspects of good HIT creation, there are other aspects that it does not cover. Among those are the qualification tasks. These tasks assess the capability of a worker before giving them access to a HIT based on the observation that each worker's skill set is different, so it is better to check their work rather than assuming that a worker can do each and every task correctly \cite{daniel2018quality}. In general a small number of golden items are given to the worker and a match to the experts allows them to go forward to work on the HITs. Qualification tasks have already been implemented in crowdsourcing platforms such as AMT and so do not need to be covered in DialCrowd 2.0.

\section{Future Work}
The DialCrowd team has connected the intent classification template of DialCrowd 2.0 to ParlAI. In this way, requesters will have access to the datasets and models ParlAI provides while having an interface to create HITs with DialCrowd 2.0. Future directions could include the community creating new templates and checking them in with ParlAI.


\section{Conclusion}
Clarity of instructions, examples, fair payment, and low quality filtering are important considerations when creating HITs so that the data gathered is of the highest quality possible. Studies have demonstrated the value of these factors. DialCrowd 2.0 puts these factors into practice by providing a set of tools that allow requesters to collect high quality data.

\section*{Acknowledgments}
This paper is supported by the National Science Foundation Graduate Research Fellowship under Grant Nos. DGE1745016 and DGE2140739. It is also partly funded by the National Science Foundation grant CNS-1512973. The opinions expressed in this paper do not necessarily reflect those of the National Science Foundation.

\section{Bibliographical References}\label{reference}

\bibliographystyle{lrec2022-bib}
\bibliography{lrec2022-example}

\begin{thebibliography}{}

\bibitem[\protect\citename{Aker \bgroup et al.\egroup }2012]{aker2012assessing}
Aker, A., El-Haj, M., Albakour, M.-D., Kruschwitz, U., et~al.
\newblock (2012).
\newblock Assessing crowdsourcing quality through objective tasks.
\newblock In {\em LREC}, pages 1456--1461. Citeseer.

\bibitem[\protect\citename{Alabduljabbar and
  Al-Dossari}2019]{alabduljabbar2019dynamic}
Alabduljabbar, R. and Al-Dossari, H.
\newblock (2019).
\newblock A dynamic selection approach for quality control mechanisms in
  crowdsourcing.
\newblock {\em IEEE Access}, 7:38644--38656.

\bibitem[\protect\citename{Allahbakhsh \bgroup et al.\egroup
  }2013]{allahbakhsh2013quality}
Allahbakhsh, M., Benatallah, B., Ignjatovic, A., Motahari-Nezhad, H.~R.,
  Bertino, E., and Dustdar, S.
\newblock (2013).
\newblock Quality control in crowdsourcing systems: Issues and directions.
\newblock {\em IEEE Internet Computing}, 17(2):76--81.

\bibitem[\protect\citename{Arkhangorodsky \bgroup et al.\egroup
  }2020]{arkhangorodsky2020meep}
Arkhangorodsky, A., Axelrod, A., Chu, C., Fang, S., Huang, Y., Nagesh, A., Shi,
  X., Zhang, B., and Knight, K.
\newblock (2020).
\newblock Meep: An open-source platform for human-human dialog collection and
  end-to-end agent training.
\newblock {\em arXiv preprint arXiv:2010.04747}.

\bibitem[\protect\citename{Buhrmester \bgroup et al.\egroup
  }2016]{buhrmester2016amazon}
Buhrmester, M., Kwang, T., and Gosling, S.~D.
\newblock (2016).
\newblock Amazon's mechanical turk: A new source of inexpensive, yet
  high-quality data?

\bibitem[\protect\citename{Chandler \bgroup et al.\egroup
  }2013]{chandler2013risks}
Chandler, J., Paolacci, G., and Mueller, P.
\newblock (2013).
\newblock Risks and rewards of crowdsourcing marketplaces.
\newblock In {\em Handbook of human computation}, pages 377--392. Springer.

\bibitem[\protect\citename{Chen \bgroup et al.\egroup
  }2011]{chen2011opportunities}
Chen, J.~J., Menezes, N.~J., Bradley, A.~D., and North, T.
\newblock (2011).
\newblock Opportunities for crowdsourcing research on amazon mechanical turk.
\newblock {\em Interfaces}, 5(3):1.

\bibitem[\protect\citename{Daniel \bgroup et al.\egroup
  }2018]{daniel2018quality}
Daniel, F., Kucherbaev, P., Cappiello, C., Benatallah, B., and Allahbakhsh, M.
\newblock (2018).
\newblock Quality control in crowdsourcing: A survey of quality attributes,
  assessment techniques, and assurance actions.
\newblock {\em ACM Computing Surveys (CSUR)}, 51(1):1--40.

\bibitem[\protect\citename{Doroudi \bgroup et al.\egroup
  }2016]{doroudi2016toward}
Doroudi, S., Kamar, E., Brunskill, E., and Horvitz, E.
\newblock (2016).
\newblock Toward a learning science for complex crowdsourcing tasks.
\newblock In {\em Proceedings of the 2016 CHI Conference on Human Factors in
  Computing Systems}, pages 2623--2634.

\bibitem[\protect\citename{Faradani \bgroup et al.\egroup }2011]{faradani2011s}
Faradani, S., Hartmann, B., and Ipeirotis, P.~G.
\newblock (2011).
\newblock What’s the right price? pricing tasks for finishing on time.
\newblock In {\em Workshops at the Twenty-Fifth AAAI Conference on Artificial
  Intelligence}.

\bibitem[\protect\citename{Feng \bgroup et al.\egroup }2009]{feng2009acquiring}
Feng, D., Besana, S., and Zajac, R.
\newblock (2009).
\newblock Acquiring high quality non-expert knowledge from on-demand workforce.
\newblock In {\em Proceedings of the 2009 Workshop on The People’s Web Meets
  NLP: Collaboratively Constructed Semantic Resources (People’s Web)}, pages
  51--56.

\bibitem[\protect\citename{Georgescu \bgroup et al.\egroup
  }2012]{georgescu2012map}
Georgescu, M., Pham, D.~D., Firan, C.~S., Nejdl, W., and Gaugaz, J.
\newblock (2012).
\newblock Map to humans and reduce error: crowdsourcing for deduplication
  applied to digital libraries.
\newblock In {\em Proceedings of the 21st ACM international conference on
  Information and knowledge management}, pages 1970--1974.

\bibitem[\protect\citename{Hsueh \bgroup et al.\egroup }2009]{hsueh2009data}
Hsueh, P.-Y., Melville, P., and Sindhwani, V.
\newblock (2009).
\newblock Data quality from crowdsourcing: a study of annotation selection
  criteria.
\newblock In {\em Proceedings of the NAACL HLT 2009 workshop on active learning
  for natural language processing}, pages 27--35.

\bibitem[\protect\citename{Huynh \bgroup et al.\egroup }2021]{huynh2021survey}
Huynh, J., Bigham, J., and Eskenazi, M.
\newblock (2021).
\newblock A survey of nlp-related crowdsourcing hits: what works and what does
  not.
\newblock {\em arXiv preprint arXiv:2111.05241}.

\bibitem[\protect\citename{K.~Chaithanya~Manam \bgroup et al.\egroup
  }2019]{k2019taskmate}
K.~Chaithanya~Manam, V., Jampani, D., Zaim, M., Wu, M.-H., and J.~Quinn, A.
\newblock (2019).
\newblock Taskmate: A mechanism to improve the quality of instructions in
  crowdsourcing.
\newblock In {\em Companion Proceedings of The 2019 World Wide Web Conference},
  pages 1121--1130.

\bibitem[\protect\citename{Kittur \bgroup et al.\egroup
  }2013]{kittur2013future}
Kittur, A., Nickerson, J.~V., Bernstein, M., Gerber, E., Shaw, A., Zimmerman,
  J., Lease, M., and Horton, J.
\newblock (2013).
\newblock The future of crowd work.
\newblock In {\em Proceedings of the 2013 conference on Computer supported
  cooperative work}, pages 1301--1318.

\bibitem[\protect\citename{Kulkarni \bgroup et al.\egroup
  }2012]{kulkarni2012mobileworks}
Kulkarni, A., Gutheim, P., Narula, P., Rolnitzky, D., Parikh, T., and Hartmann,
  B.
\newblock (2012).
\newblock Mobileworks: Designing for quality in a managed crowdsourcing
  architecture.
\newblock {\em IEEE Internet Computing}, 16(5):28--35.

\bibitem[\protect\citename{Lee \bgroup et al.\egroup }2018]{lee2018dialcrowd}
Lee, K., Zhao, T., Black, A.~W., and Eskenazi, M.
\newblock (2018).
\newblock Dialcrowd: A toolkit for easy dialog system assessment.
\newblock In {\em Proceedings of the 19th Annual SIGdial Meeting on Discourse
  and Dialogue}, pages 245--248.

\bibitem[\protect\citename{Marcus \bgroup et al.\egroup
  }2012]{marcus2012counting}
Marcus, A., Karger, D., Madden, S., Miller, R., and Oh, S.
\newblock (2012).
\newblock Counting with the crowd.
\newblock {\em Proceedings of the VLDB Endowment}, 6(2):109--120.

\bibitem[\protect\citename{Mason and Watts}2009]{mason2009financial}
Mason, W. and Watts, D.~J.
\newblock (2009).
\newblock Financial incentives and the" performance of crowds".
\newblock In {\em Proceedings of the ACM SIGKDD workshop on human computation},
  pages 77--85.

\bibitem[\protect\citename{Miller \bgroup et al.\egroup
  }2017]{miller2017parlai}
Miller, A.~H., Feng, W., Fisch, A., Lu, J., Batra, D., Bordes, A., Parikh, D.,
  and Weston, J.
\newblock (2017).
\newblock Parlai: A dialog research software platform.
\newblock {\em arXiv preprint arXiv:1705.06476}.

\bibitem[\protect\citename{Mortensen \bgroup et al.\egroup
  }2017]{mortensen2017exploration}
Mortensen, M.~L., Adam, G.~P., Trikalinos, T.~A., Kraska, T., and Wallace,
  B.~C.
\newblock (2017).
\newblock An exploration of crowdsourcing citation screening for systematic
  reviews.
\newblock {\em Research synthesis methods}, 8(3):366--386.

\bibitem[\protect\citename{Nielsen}1994]{nielsen1994usability}
Nielsen, J.
\newblock (1994).
\newblock {\em Usability engineering}.
\newblock Morgan Kaufmann.

\bibitem[\protect\citename{Nouri \bgroup et al.\egroup }2021]{nouri2021unclear}
Nouri, Z., Gadiraju, U., Engels, G., and Wachsmuth, H.
\newblock (2021).
\newblock What is unclear? computational assessment of task clarity in
  crowdsourcing.
\newblock In {\em Proceedings of the 32nd ACM Conference on Hypertext and
  Social Media}, pages 165--175.

\bibitem[\protect\citename{Paolacci \bgroup et al.\egroup
  }2010]{paolacci2010running}
Paolacci, G., Chandler, J., and Ipeirotis, P.~G.
\newblock (2010).
\newblock Running experiments on amazon mechanical turk.
\newblock {\em Judgment and Decision making}, 5(5):411--419.

\bibitem[\protect\citename{Rzeszotarski and
  Kittur}2012]{rzeszotarski2012crowdscape}
Rzeszotarski, J. and Kittur, A.
\newblock (2012).
\newblock Crowdscape: interactively visualizing user behavior and output.
\newblock In {\em Proceedings of the 25th annual ACM symposium on User
  interface software and technology}, pages 55--62.

\bibitem[\protect\citename{Sayeed \bgroup et al.\egroup
  }2011]{sayeed2011crowdsourcing}
Sayeed, A., Rusk, B., Petrov, M., Nguyen, H.~C., Meyer, T.~J., and Weinberg, A.
\newblock (2011).
\newblock Crowdsourcing syntactic relatedness judgements for opinion mining in
  the study of information technology adoption.
\newblock In {\em proceedings of the 5th ACL-HLT workshop on language
  technology for cultural heritage, social sciences, and humanities}, pages
  69--77.

\bibitem[\protect\citename{Wang \bgroup et al.\egroup }2011]{wang2011managing}
Wang, J., Ipeirotis, P.~G., and Provost, F.
\newblock (2011).
\newblock Managing crowdsourcing workers.
\newblock In {\em The 2011 winter conference on business intelligence}, pages
  10--12. Citeseer.

\bibitem[\protect\citename{Wang \bgroup et al.\egroup }2017]{wang2017cost}
Wang, J., Ipeirotis, P.~G., and Provost, F.
\newblock (2017).
\newblock Cost-effective quality assurance in crowd labeling.
\newblock {\em Information Systems Research}, 28(1):137--158.

\bibitem[\protect\citename{Zhu \bgroup et al.\egroup
  }2020]{zhu-etal-2020-convlab}
Zhu, Q., Zhang, Z., Fang, Y., Li, X., Takanobu, R., Li, J., Peng, B., Gao, J.,
  Zhu, X., and Huang, M.
\newblock (2020).
\newblock {C}onv{L}ab-2: An open-source toolkit for building, evaluating, and
  diagnosing dialogue systems.
\newblock In {\em Proceedings of the 58th Annual Meeting of the Association for
  Computational Linguistics: System Demonstrations}, pages 142--149, Online,
  July. Association for Computational Linguistics.

\end{thebibliography}


\newpage
\section*{Appendix}
Figures 1, 2, and 3 are from the configuration page of DialCrowd 2.0. Figures 4 and 5 are from the task page that the workers see.

\begin{figure*}[htp]
    \centering
    \includegraphics[width=\textwidth]{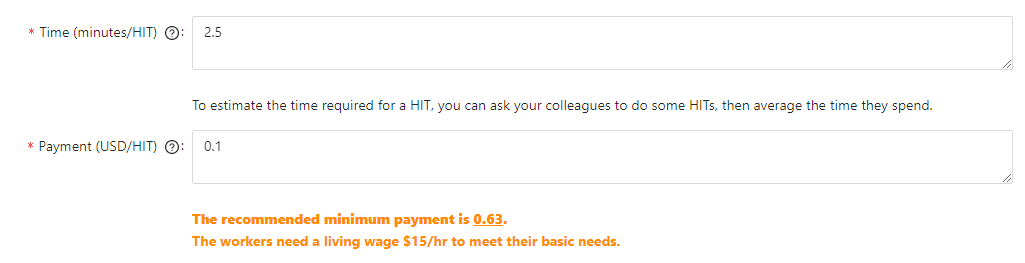}
    \caption{DialCrowd 2.0 will calculate and suggest a minimum payment for the HIT based on the time estimate scaled to \$15/hr}
\end{figure*}


\begin{figure*}[htp]
    \centering
    \includegraphics[width=\textwidth]{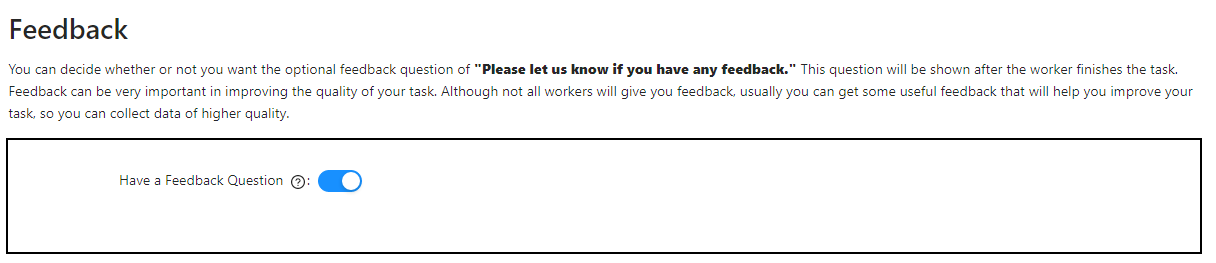}
    \caption{Feedback Option for the Requesters}
\end{figure*}

\begin{figure*}[htp]
    \centering
    \includegraphics[width=\textwidth]{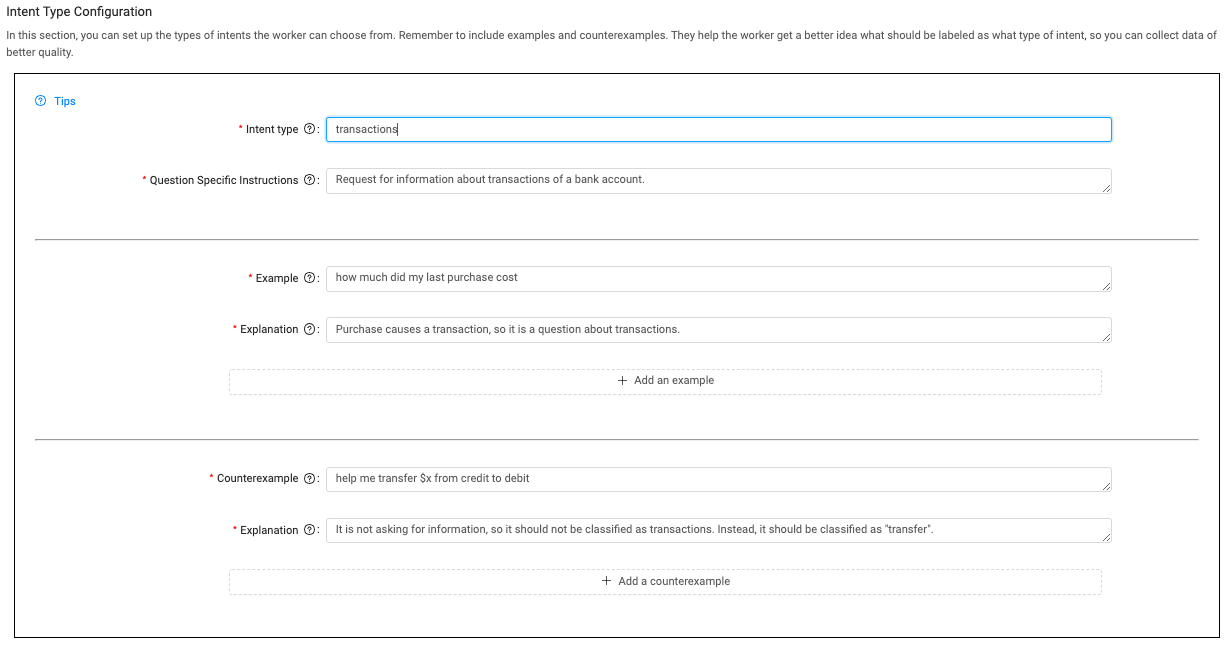}
    \caption{Using Examples and Counterexamples For Specific Intents}
\end{figure*}

\begin{figure*}[htp]
    \centering
    \includegraphics[width=\textwidth]{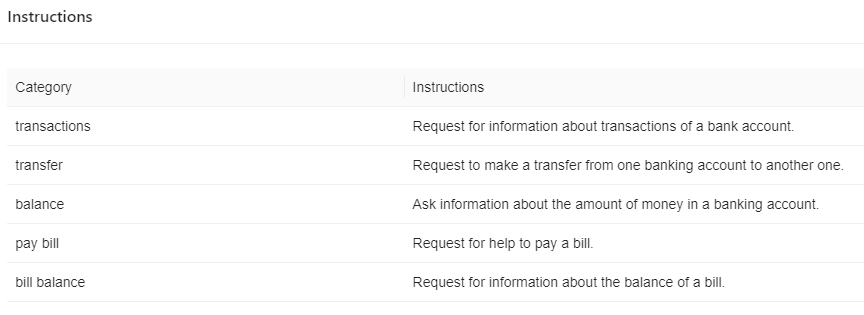}
    \caption{Instructions For Specific Intents}
\end{figure*}

\begin{figure*}[htp]
    \centering
    \includegraphics[width=\textwidth]{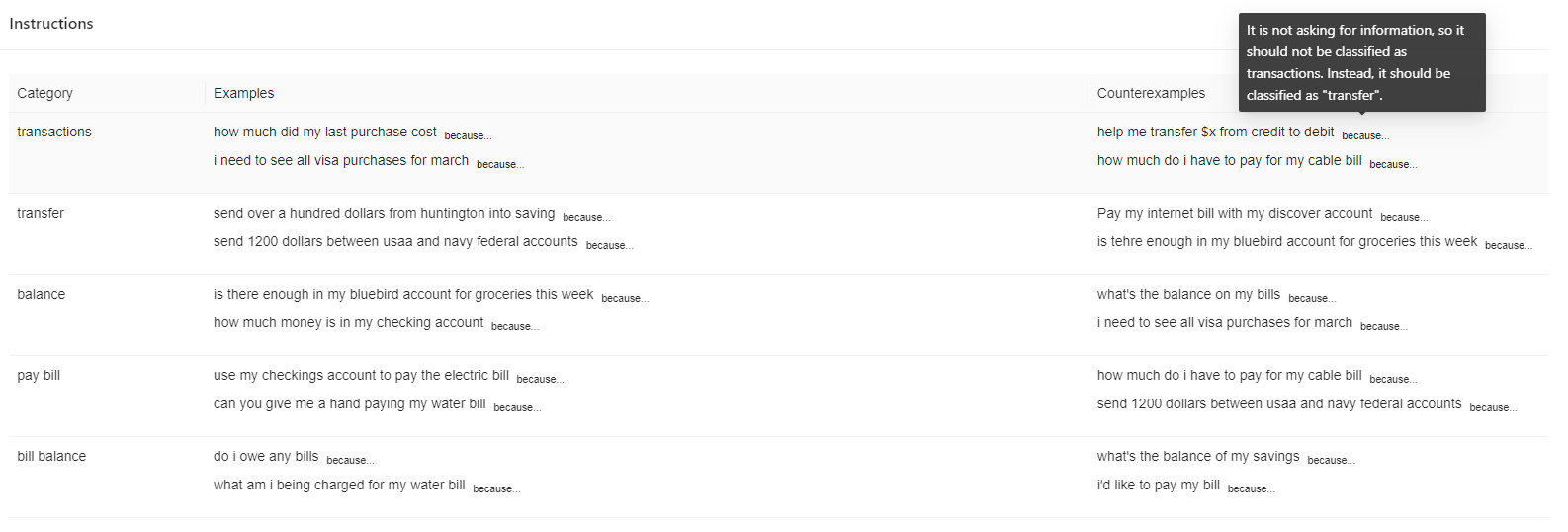}
    \caption{Examples For Specific Intents}
\end{figure*}

\end{document}